\documentclass{bmvc2k}

\usepackage{amssymb}
 \usepackage{dsfont}
\usepackage[table,xcdraw]{xcolor}
\usepackage{arydshln}

\title{Sensor-Independent Illumination Estimation for DNN Models}

\addauthor{Mahmoud Afifi$^1$}{http://www.cse.yorku.ca/~mafifi/}{1}
\addauthor{Michael S. Brown$^{1,2}$}{http://www.cse.yorku.ca/~mbrown/}{2}

\addinstitution{
Lassonde School of Engineering$^1$\\
York University\\
Toronto, Canada
}
\addinstitution{
Samsung AI Center (SAIC)$^2$\\
Samsung Research\\
 Toronto, Canada
}

\runninghead{Afifi and Brown}{Sensor-Independent Illumination Estimation}

\begin{document}

\maketitle

\begin{abstract}
While modern deep neural networks (DNNs) achieve state-of-the-art results for illuminant estimation, it is currently necessary to train a separate DNN for each type of camera sensor. This means when a camera manufacturer uses a new sensor, it is necessary to retrain an existing DNN model with training images captured by the new sensor. This paper addresses this problem by introducing a novel sensor-independent illuminant estimation framework. Our method learns a sensor-independent {\it working space} that can be used to canonicalize the RGB values of any arbitrary camera sensor. Our learned space retains the linear property of the original sensor raw-RGB space and allows unseen camera sensors to be used on a single DNN model trained on this working space.  We demonstrate the effectiveness of this approach on several different camera sensors and show it provides performance on par with state-of-the-art methods that were trained per sensor.
\end{abstract}

\section{Introduction and Motivation}
\label{sec:intro}
Color constancy is the constant appearance of object colors under different illumination conditions~\cite{foster2003does}. Human vision has this illumination adaption ability to recognize the same object colors under different scene lighting~\cite{maloney1999physics}.
Camera sensors, however, do not have this ability and as a result, computational color constancy is required to be applied onboard the camera. In a photography context, this procedure is typically termed {\it white balance}.  The key challenge for computational color constancy is the ability to estimate a camera sensor's RGB response to the scene's illumination. Illumination estimation, or auto white balance (AWB), is a fundamental procedure applied onboard all cameras and is critical in ensuring the correct interpretation of scene colors.

Computational color constancy can be described in terms of the physical image formation process as follows. Let $\mathbf{I} = \{\mathbf{I}_r, \mathbf{I}_g, \mathbf{I}_b\}$ denote an image captured in the linear raw-RGB space. The value of each color channel $c = \{\text{R}, \text{G}, \text{B}\}$ for a pixel located at $x$ in $\mathbf{I}$ is given by the following equation \cite{basri2003lambertian}:
\begin{equation}
\label{eq0}
\mathbf{I}_c(x) =\int_{\gamma} \rho(x,\lambda)R(x,\lambda)S_{c}(\lambda) d\lambda,
\end{equation}
\noindent
where $\gamma$ is the visible light spectrum (approximately 380nm to 780nm), $\rho(\cdot)$ is the illuminant spectral power distribution, $R(\cdot)$ is the captured scene's spectral reflectance properties, and $S(\cdot)$ is the camera sensor response function at wavelength $\lambda$. The problem can be simplified by assuming a single uniform illuminant in the scene as follows:
\begin{equation}
\label{eq1}
\mathbf{I}_c =  \pmb{\ell}_{c}\!\times\!\mathbf{R}_{c},
\end{equation}
\noindent
where $\pmb{\ell}_{c}$ is the scene illuminant value of color channel $c$. A standard approach to this problem is to use a linear model (i.e., a $3\!\times\!3$ diagonal matrix) such that $\pmb{\ell}_{\text{R}} = \pmb{\ell}_{\text{G}} = \pmb{\ell}_{\text{B}}$ (i.e., white illuminant).

\begin{figure}
\includegraphics[width=\linewidth]{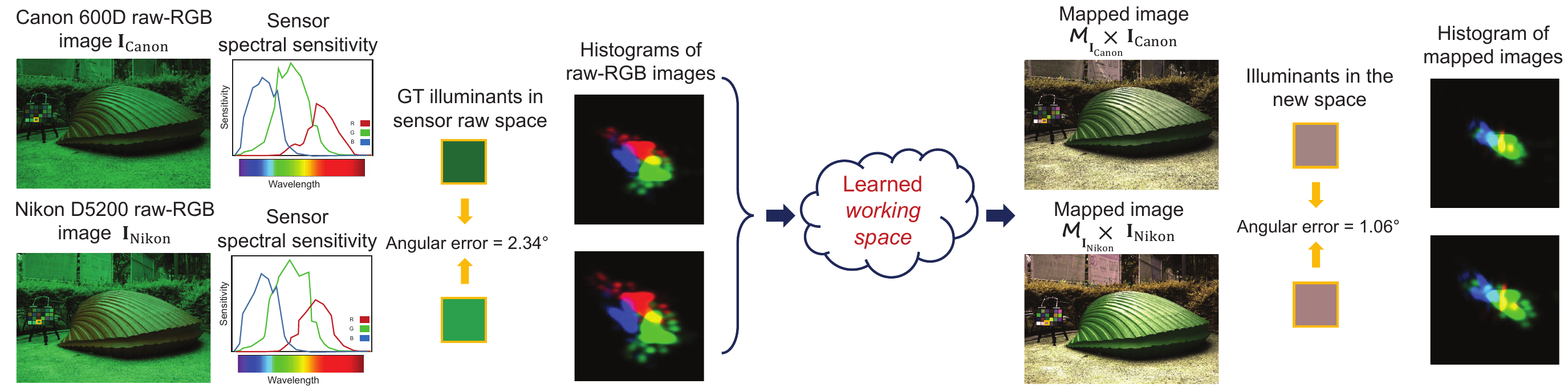}
\vspace{-5mm}
\caption{A scene captured by two different camera sensors results in different ground truth illuminants due to different camera sensor responses. We learn a device-independent {\it working space} that reduces the difference between ground truth illuminants of the same scenes.}
\label{fig:fig1}\vspace{-5mm}
\end{figure}

Typically, $\pmb{\ell}$ is unknown and should be defined to obtain the true objects' body reflectance values $\mathbf{R}$ in the input image $\mathbf{I}$. The value of $\pmb{\ell}$ is specific to the camera sensor response function $S(\cdot)$, meaning that the same scene captured by different camera sensors results in different values of $\pmb{\ell}$. Fig. \ref{fig:fig1} shows an example.

Illuminant estimation methods aim to estimate the value $\pmb{\ell}$ from the sensor's raw-RGB image. Recently, deep neural network (DNN) methods have demonstrated state-of-the-art results for the illuminant estimation task. These approaches, however, need to train the DNN model per camera sensor. This is a significant drawback. When a camera manufacturer decides to use a new sensor, the DNN model will need to be retrained on a new image dataset captured by the new sensor. Collecting such datasets with the corresponding ground-truth illuminant raw-RGB values is a tedious process. As a result, many AWB algorithms deployed on cameras still rely on simple statistical-based methods even though the accuracy is not comparable to those obtained by the learning-based methods~\cite{Afifi19}.

\begin{figure}
\includegraphics[width=\linewidth]{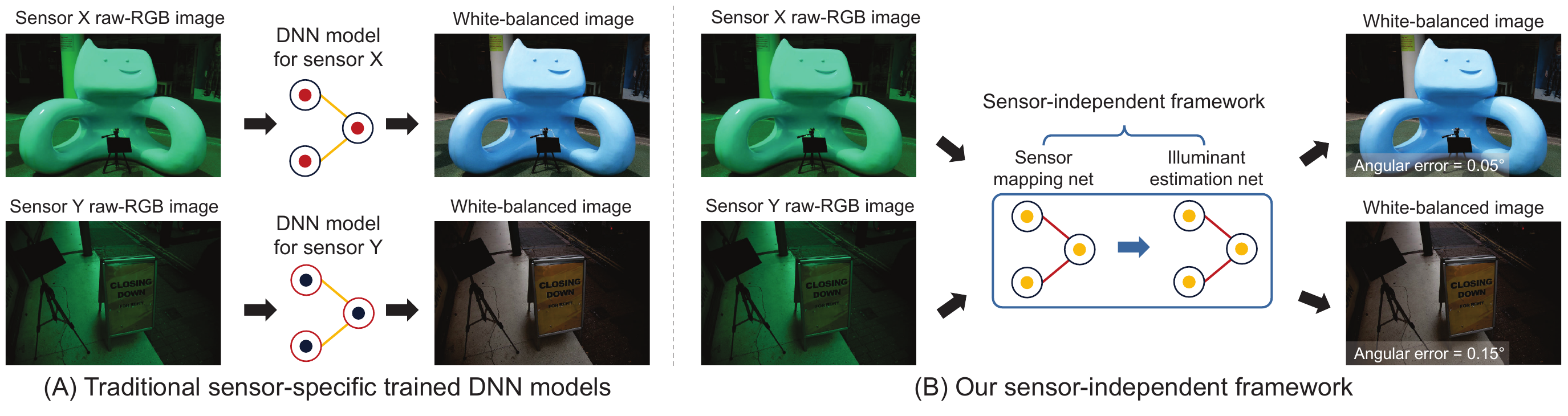}
\vspace{-5mm}
\caption{(A) Traditional learning-based illuminant estimation methods train or fine-tune a model per camera sensor. (B) Our method can be trained on images captured by different camera sensors and generalizes well for unseen camera sensors. Shown images are rendered in the sRGB color space by the camera imaging pipeline in \cite{karaimer2016software} to aid visualization.}
\label{fig:teaser}\vspace{-5mm}
\end{figure}

\paragraph{Contribution}~In this paper, we introduce a sensor-independent learning framework for illuminant estimation. The idea is similar to the color space conversion process applied onboard cameras that maps the sensor-specific RGB values to a perceptual-based color space -- namely, CIE XYZ. The color space conversion process estimates a color space transform (CST) matrix to map white-balanced sensor-specific raw-RGB images to CIE XYZ~\cite{ramanath2005color, karaimer2016software}.  This process is applied onboard cameras {\it after} the illuminant estimation and white-balance step, and relies on the estimated scene illuminant to compute the CST matrix~\cite{can2018improving}. Our solution, however, is to learn a new space that is used \textit{before} the illuminant estimation step. Specifically, we design a novel unsupervised deep learning framework that learns how to map each input image, captured by arbitrary camera sensor, to a non-perceptual sensor-independent {\it working space}. Mapping input images to this space allows us to train our model using training sets captured by different camera sensors, achieving good accuracy and generalizing well for unseen camera sensors, as shown in Fig. \ref{fig:teaser}.

\section{Related Work}
\label{sec:relted}

We discuss two areas of related work: (i) illumination estimation and (ii) mapping camera sensor responses.

\subsection{Illumination Estimation}
As previously discussed, illuminant estimation is the key routine that makes up a camera's AWB function.  Illuminant estimation methods aim to estimate the illumination in the imaged scene directly from a raw-RGB image without a known achromatic reference scene patch. We categorize the illuminant estimation methods into two categories, which are: (i) sensor-independent methods and (ii) sensor-dependent methods.

\paragraph{Sensor-Independent Methods}These methods operate using statistics from an image's color distribution and spatial layout to estimate the scene illuminant. Representative statistical-based methods include: Gray-World \cite{GW}, White-Patch \cite{maxRGB}, Shades-of-Gray \cite{SoG}, Gray-Edges, and PCA-based Bright-and-Dark Colors \cite{cheng2014illuminant}. These methods are fast and easy to implement; however, their results are not always satisfactory.

\paragraph{Sensor-Dependent Methods}Learning-based models outperform statistical-based methods by training sensor-specific models on training examples provided with the labeled images with ground-truth illumination obtained from physical charts placed in the scene with achromatic reference patches.   These training images are captured with the sensor make and model being trained. Representative examples include Bayesian-based methods \cite{brainard1997bayesian, rosenberg2004bayesian, gehler2008bayesian}, gamut-based methods \cite{forsyth1990novel, Gamut, PixelGamut}, exemplar-based methods \cite{gijsenij2011color,ExemplarCC, banic2015color}, bias-correction methods \cite{MomentCorrection, royalsociety, Afifi19}, and, more recently, DNN methods \cite{BMVC1, barron2015convolutional, Seoung, shi2016deep,  hu2017fc, barron2017fast}, including few-shot learning \cite{mcdonagh2018meta}. The obvious drawback of these methods is that they do not generalize well for arbitrary camera sensors without retraining/fine-tuning on samples captured by testing camera sensor. Our learning method, however, is explicitly designed to be sensor-independent and generalizes well for unseen camera sensors without the need to retrain/tune our model.

\subsection{Mapping Camera Sensor Responses}
Another research topic related to our work is mapping camera raw-RGB sensor responses to a perceptual color space. This process is applied onboard digital cameras to map the captured sensor-specific raw-RGB image to a standard device-independent ``canonical'' space (e.g., CIE XYZ) \cite{ramanath2005color, karaimer2016software}. Usually this conversion is performed using a $3\!\times\!3$ matrix and requires an accurate estimation of the scene illuminant \cite{can2018improving}.  It is important to note that this mapping to CIE XYZ requires that white-balance procedure first be applied.  As a result, it is not possible to use CIE XYZ as the canonical color space to perform illumination estimation.

Work by Nguyen~et al. \cite{nguyen2014raw} studied several transformations to map responses from a source camera sensor to a target camera sensor, instead of mapping to a perceptual space. In their study, a color rendition reference chart is captured by both source and target camera sensors in order to compute the raw-to-raw mapping function. Learning a mapping transformation between responses of two different sensors is also adapted in \cite{gao2017improving}. While our approach is similar in this goal, the work in~\cite{nguyen2014raw, gao2017improving} has no mechanism to map an unseen sensor to a canonical working space without explicit calibration.

\section{Proposed Method}
\label{sec:method}
\begin{figure}
\includegraphics[width=\linewidth]{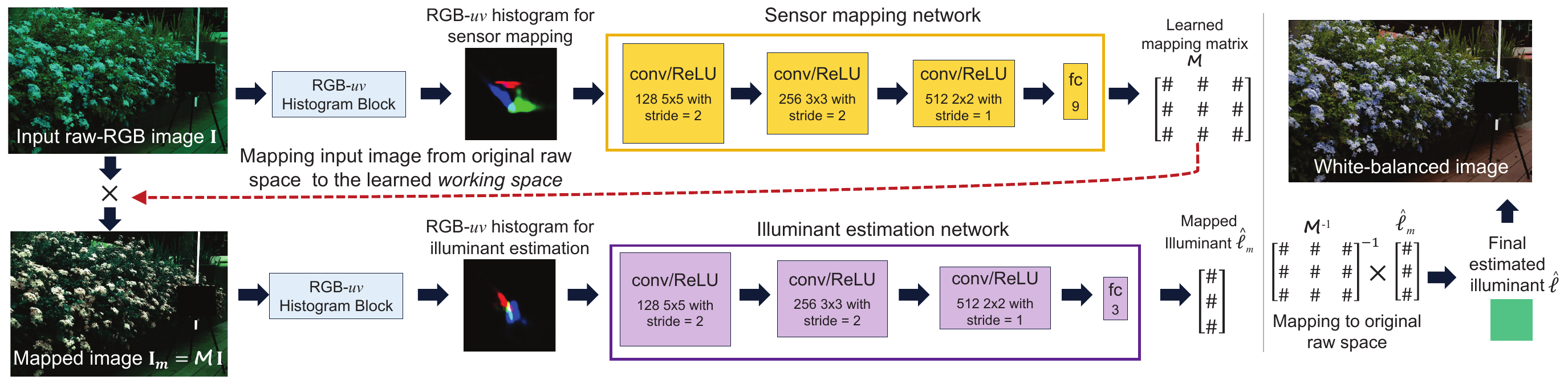}
\vspace{-5mm}
\caption{Our proposed method consists of two networks: (i) a sensor mapping network and (ii) an illuminant estimation network. Our networks are trained jointly in an end-to-end manner to learn an image-specific mapping matrix (resulting from the sensor mapping network) and scene illuminant in the learned space (resulting from the illuminant estimation network). The final estimated illuminant is produced by mapping the result illuminant from our learned space to the input image's camera-specific raw space.}
\label{fig:main}\vspace{-5mm}
\end{figure}

Fig.~\ref{fig:main} provides an overview of our framework. Our method accepts thumbnail ($150\!\times\!150$ pixels) linear raw-RGB images, captured by an arbitrary camera sensor, and estimates scene illuminant RGB vectors in the same space of input images.

We rely on color distribution of input thumbnail image $\mathbf{I}$ to estimate an image-specific transformation matrix that maps the input image to our working space. This mapping allows us to accept images captured by different sensors and estimate scene illuminant values in the original space of input images.

We begin with the formulation of our problem followed by a detailed description of our framework components and the training process. Note that we will assume input raw-RGB images are represented as $3\!\times\!n$ matrices, where $n=150\!\times\!150$ is the total number of pixels in the thumbnail image and the three rows represent the R, G, and B values.

\subsection{Problem Formulation}
\label{subsec:overview}
We propose to work in a new learned space for illumination estimation. This space is sensor-independent and retains the linear property of the original raw-RGB space. To that end, we introduce a learnable $3\!\times\!3$ matrix $\pmb{\mathcal{M}}$ that maps an input image $\mathbf{I}$ from its original sensor-specific space to a new working space. We can reformulate Eq. \ref{eq1} as follows:
\begin{equation}
\label{eq2}
\pmb{\mathcal{M}}^{-1}\pmb{\mathcal{M}}\mathbf{I} =  \texttt{diag}(\pmb{\mathcal{M}}^{-1}\pmb{\mathcal{M}}\pmb{\ell}) \mathbf{R},
\end{equation}
\noindent
where $\texttt{diag}(\cdot)$ is a diagonal matrix and  $\pmb{\mathcal{M}}$ is a learned matrix that maps arbitrary sensor responses to a sensor-independent space.

Given a mapped image $\mathbf{I}_m = \pmb{\mathcal{M}}\mathbf{I}$ in our learned space, we aim to estimate the mapped vector $\pmb{\ell}_m = \pmb{\mathcal{M}}\pmb{\ell}$ that represents the scene illumination values of $\mathbf{I}_m$ in the new space. The original scene illuminant (represented in the original sensor raw-RGB space) can be reconstructed by the following equation:
\begin{equation}
\label{eq3}
\pmb{\ell} = \pmb{\mathcal{M}}^{-1}\pmb{\ell}_m.
\end{equation}

\subsection{RGB--$uv$ Histogram Block}
\label{subsec:RGB-uvHist}
Prior work has shown that the illumination estimation problem is related primarily to the image's color distribution~\cite{cheng2014illuminant, barron2015convolutional}.  Accordingly, we use the image's color distribution as an input for our method. Representing the image using a full 3D RGB histogram requires significant amounts of memory -- for example, a $256^3$ RGB histogram requires more than 16 million entries. Down-sampling the histogram -- for example, to 64-bins -- still requires a considerable amount of memory.

Our method relies on the RGB-$uv$ histogram feature used in \cite{afifi2019color}. This feature represents the image color distribution in the log of chromaticity space \cite{drew2003recovery}. Unlike the original RGB-$uv$ feature, we use two learnable parameters to control the contribution of each color channel in the generated histogram and the smoothness of histogram bins. Specifically, the RGB-$uv$ histogram block represents the color distribution of an image $\mathbf{I}$ as a three-layer histogram $\mathbf{H}(\mathbf{I})$  represented as an $m\!\times\!m\!\times 3$ tensor. The produced histogram $\mathbf{H}(\mathbf{I})$ is parameterized by $uv$ and computed as follows:
\begin{equation}
\label{eq2_UVHist}
\begin{gathered}
\mathbf{I}_y(i) = \sqrt{\mathbf{I}_{\textrm{R}(i)}^{2} + \mathbf{I}_{\textrm{G}(i)}^{2} + \mathbf{I}_{\textrm{B}(i)}^{2}},
\\
\mathbf{I}_{u1(i)} = \log{\left(\frac{ \mathbf{I}_{\textrm{R}(i)}}{\mathbf{I}_{\textrm{G}(i)}} + \epsilon\right)} \textrm{ , }  \mathbf{I}_{v1(i)} = \log{\left(\frac{ \mathbf{I}_{\textrm{R}(i)}}{\mathbf{I}_{\textrm{B}(i)}}+ \epsilon\right)},
\\
\mathbf{I}_{u2(i)} = \log{\left(\frac{ \mathbf{I}_{\textrm{G}(i)}}{\mathbf{I}_{\textrm{R}(i)}}+ \epsilon\right)} \textrm{ , }  \mathbf{I}_{v2(i)} = \log{\left(\frac{ \mathbf{I}_{\textrm{G}(i)}}{\mathbf{I}_{\textrm{B}(i)}}+ \epsilon\right)},
\\
\mathbf{I}_{u3(i)} = \log{\left(\frac{ \mathbf{I}_{\textrm{B}(i)}}{\mathbf{I}_{\textrm{R}(i)}}+ \epsilon\right)} \textrm{ , }  \mathbf{I}_{v3(i)} = \log{\left(\frac{ \mathbf{I}_{\textrm{B}(i)}}{\mathbf{I}_{\textrm{G}(i)}}+ \epsilon\right)},
\\
\mathbf{H}\left(\mathbf{I}\right)_{(u,v,c)} = \left(s_{c} \sum_{i} \mathbf{I}_{y(i)}  \exp{\left(-\left| \mathbf{I}_{uc(i)} - u \right|/\sigma_c^2\right)}   \exp{\left(-\left| \mathbf{I}_{vc(i)} - v \right|/\sigma_c^2\right)}\right)^{1/2},
\end{gathered}
\end{equation}
\noindent
where  $i = \{1,...,n\}$, $c \in {\{1, 2, 3\}}$ represents each color channel in $\mathbf{H}$, $\epsilon$ is a small positive constant added for numerical stability, and $s_{c}$ and $\sigma_c$ are learnable scale and fall-off parameters, respectively. The scale factor $s_{c}$ controls the contribution of each layer in our histogram, while the fall-off factor $\sigma_c$ controls the smoothness of the histogram's bins of each layer. The values of these parameters (i.e., $s_{c}$ and $\sigma_c$) are learned during the training phase.

\subsection{Network Architecture}
\label{subsec:net}

As shown in Fig. \ref{fig:main}, our framework consists of two networks: (i) a sensor mapping network and (ii) an illuminant estimation network. The input to each network is the RGB-$uv$ histogram feature produced by our histogram block. The sensor mapping network accepts an RGB-$uv$ histogram of a thumbnail raw-RGB image $\mathbf{I}$ in its original sensor space, while the illuminant estimation network accepts RGB-$uv$ histograms of the mapped image $\mathbf{I}_m$ to our learned space. In our implementation, we use $m=61$ and each histogram feature is represented by a $61\!\times\!61\!\times\!3$ tensor.

We use a simple network architecture for each network. Specifically, each network consists of three conv/ReLU layers followed by a fully connected (fc) layer. The kernel size and stride step used in each conv layer are shown in Fig. \ref{fig:main}.

In the sensor mapping network, the last fc layer has nine neurons. The output vector $\pmb{v}$ of this fc layer is reshaped to construct a $3\!\times\!3$ matrix $\pmb{V}$, which is used to build $\pmb{\mathcal{M}}$ as described in the following equation:
\begin{equation}
\pmb{\mathcal{M}} =  \frac{1}{\lVert\pmb{V}\rVert_1 + \epsilon} \left|\pmb{V}\right|,
\end{equation}
\noindent where $\left|\cdot\right|$ is the modulus (absolute magnitude), $\lVert\cdot\rVert_1$ is the matrix 1-norm, and  $\epsilon$ is added for numerical stability. The modulus step is necessary to avoid negative values in the mapped image $\mathbf{I}_m$, while the normalization step is used to avoid having extremely large values in $\mathbf{I}_m$. Note the values of $\pmb{\mathcal{M}}$ are image-specific, meaning that its values are produced based on the input image's color distribution in the original raw-RGB space.

There are three neurons in the last fc layer of the illuminant estimation network to produce illuminant vector $\hat{\pmb{\ell}}_m$ of the mapped image $\mathbf{I}_m$. Note that the estimated vector $\hat{\pmb{\ell}}_m$ represents the scene illuminant in our learned space. The final result is obtained by mapping $\hat{\pmb{\ell}}_m$ back to the original space of $\mathbf{I}$ using Eq. \ref{eq3}.

\subsection{Training}
\label{subsec:training}
We jointly train our sensor mapping and illuminant estimation networks in an end-to-end manner using the adaptive moment estimation (Adam) optimizer \cite{kingma2014adam} with a decay rate of gradient moving average $\beta_1=0.85$, a decay rate of squared gradient moving average $\beta_2=0.99$, and a mini-batch with eight observations at each iteration. We initialized both network weights with Xavier initialization \cite{glorot2010understanding}. The learning rate was set to $10^{-5}$ and decayed every five epochs.

We adopt the recovery angular error (referred to as the angular error) as our loss function \cite{finlayson1995color}. The angular error is computed between the ground truth illuminant $\pmb{\ell}$ and our estimated illuminant $\hat{\pmb{\ell}}_m$ after mapping it to the original raw-RGB space of training image $\mathbf{I}$. The loss function can be described by the following equation:
\begin{equation}
L(\hat{\pmb{\ell}}_m, \pmb{\mathcal{M}}) = \cos^{-1}\left( \frac{ \pmb{\ell} \cdot \left(\pmb{\mathcal{M}}^{-1} \hat{\pmb{\ell}}_m\right)}{\lVert \pmb{\ell} \rVert \lVert  \pmb{\mathcal{M}}^{-1} \hat{\pmb{\ell}}_m\rVert}\right),
\end{equation}
where $\lVert\cdot\rVert$ is the Euclidean norm, and ($\cdot$) is the vector dot-product.

As the values of $\pmb{\mathcal{M}}$ are produced by the sensor mapping network, there is a possibility of producing a singular matrix output. In this case, we add small  offset $\mathcal{N}(0,1)\!\times\!10^{-4}$ to each parameter in $\pmb{\mathcal{M}}$ to make it invertible.

\begin{figure}
\includegraphics[width=\linewidth]{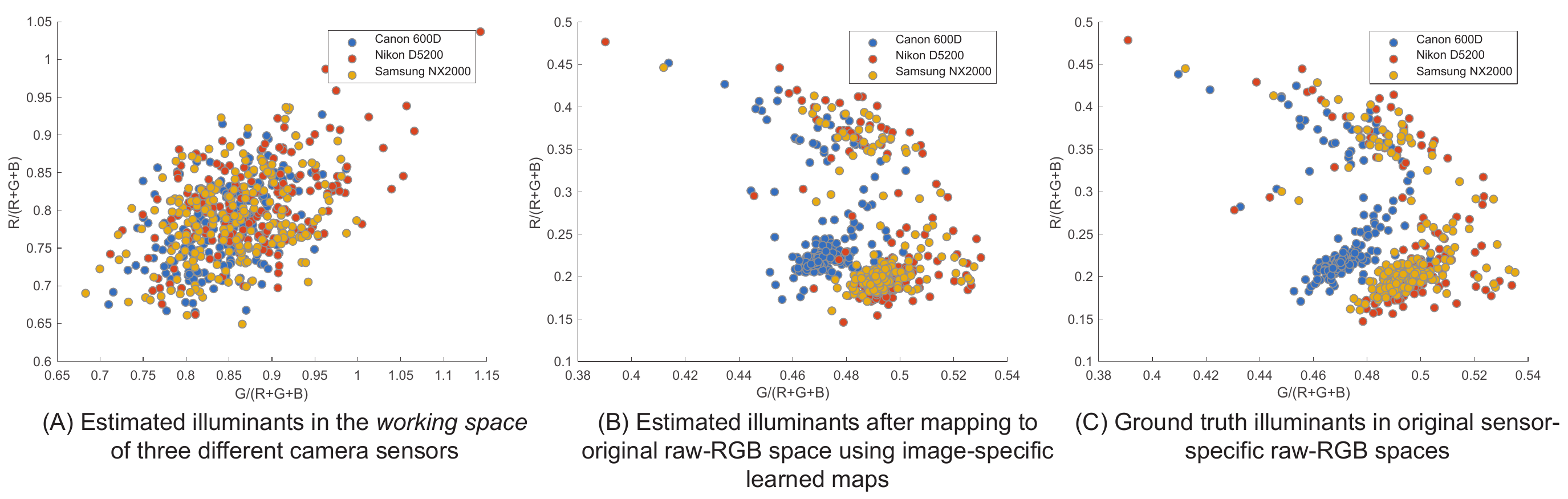}
\vspace{-5mm}
\caption{Raw-RGB images capture the same set of scenes using three different cameras taken from the NUS 8-Cameras dataset \cite{cheng2014illuminant}. (A) Estimated illuminants resulted from the illuminant estimation network in our learned {\it working space}. (B) Estimated illuminants after mapping to the original raw-RGB space. This mapping is performed by multiplying each illuminant vector by the inverse of the learned image-specific mapping matrix (resulting from the sensor mapping network). (C) Corresponding ground truth illuminants in the original raw-RGB space of each image.}
\label{fig:points}\vspace{-5mm}
\end{figure}

At the end of the training process, our framework learns an image-specific matrix $\pmb{\mathcal{M}}$ that maps an input image taken by an arbitrary sensor to the learned space. Fig. \ref{fig:points} shows an example of three different camera responses capturing the same set of scenes. As shown in Fig. \ref{fig:points}-(A), the estimated illuminants of these sensors are bounded in the learned space. These illuminants are mapped back to the original raw-RGB sensor space of the corresponding input images using Eq. \ref{eq3}. As shown in Fig. \ref{fig:points}-(B) and Fig. \ref{fig:points}-(C), our final estimated illuminants are close to the ground truth illuminants of each camera sensor.

\section{Experimental Results}
\label{sec:results}

\begin{table}
\caption{Angular errors on the NUS 8-Cameras \cite{cheng2014illuminant} and Gehler-Shi \cite{gehler2008bayesian} datasets. Methods highlighted in gray are trained/tuned for each camera sensor (i.e., sensor-specific models). The lowest errors are highlighted in yellow. \label{Table1}}\vspace{2mm}
\parbox{.48\linewidth}{
\centering
\scalebox{0.578}{
\begin{tabular}{l|cccc}
\textbf{\begin{tabular}[c]{@{}l@{}}\textbf{NUS 8-Cameras Dataset}\\\textbf{Method} \end{tabular}}& \textbf{Mean} & \textbf{Med.} & \textbf{\begin{tabular}[c]{@{}l@{}}Best \\ 25\%\end{tabular}} & \textbf{\begin{tabular}[c]{@{}l@{}}Worst \\ 25\%\end{tabular}} \\ \hline
White-Patch \cite{maxRGB} & 9.91 & 7.44 & 1.44 & 21.27 \\
Pixel-based Gamut \cite{PixelGamut} & 5.27 & 4.26 & 1.28 & 11.16 \\
Grey-world (GW) \cite{GW} & 4.59 & 3.46 & 1.16 & 9.85 \\
Edge-based Gamut \cite{PixelGamut}& 4.40 & 3.30 & 0.99 & 9.83 \\
Shades-of-Gray \cite{SoG}& 3.67 & 2.94 & 0.98 & 7.75 \\
\cellcolor[HTML]{EFEFEF}Bayesian \cite{gehler2008bayesian} & 3.50 & 2.36 & 0.78 & 8.02 \\
Local Surface Reflectance \cite{gao2014efficient}& 3.45 & 2.51 & 0.98 & 7.32 \\
2nd-order Gray-Edge \cite{GE} & 3.36 & 2.70 & 0.89 & 7.14 \\
1st-order Gray-Edge \cite{GE}& 3.35 & 2.58 & 0.79 & 7.18 \\
Quasi-unsupervised \cite{bianco2019quasi} & 3.00 & 2.25 & - & - \\
\cellcolor[HTML]{EFEFEF}Corrected-Moment \cite{MomentCorrection} & 2.95 & 2.05 & 0.59 & 6.89 \\
PCA-based B/W Colors \cite{cheng2014illuminant}& 2.93 & 2.33 & 0.78 & 6.13 \\
Grayness Index \cite{GI}& 2.91 & 1.97 & 0.56 & 6.67 \\

\cellcolor[HTML]{EFEFEF}Color Dog \cite{banic2015color} & 2.83 & 1.77 & 0.48 & 7.04 \\
\cellcolor[HTML]{EFEFEF}APAP using GW \cite{Afifi19}& 2.40 & 1.76 & 0.55 & 5.42 \\
\cellcolor[HTML]{EFEFEF}Conv Color Constancy \cite{barron2015convolutional}& 2.38 & 1.69 & 0.45 & 5.85 \\
\cellcolor[HTML]{EFEFEF}Effective Regression Tree \cite{cheng2015effective}& 2.36 &  1.59 & 0.49 & 5.54 \\
\cellcolor[HTML]{EFEFEF}WB-sRGB (modified for raw-RGB) \cite{afifi2019color}& 2.26 & 1.60 & 0.48 &
5.21 \\
\cellcolor[HTML]{EFEFEF}Deep Specialized Net \cite{shi2016deep}& 2.24 & 1.46 & 0.48 & 6.08 \\
\cellcolor[HTML]{EFEFEF}Meta-AWB w 20 tuning images \cite{mcdonagh2018meta}& 2.23 & 1.49 & 0.49 & 5.20 \\
\cellcolor[HTML]{EFEFEF}SqueezeNet-FC4 & 2.23 & 1.57 & 0.47 & 5.15 \\
\cellcolor[HTML]{EFEFEF}AlexNet-FC4 \cite{hu2017fc}& 2.12 & 1.53 & 0.48 & 4.78 \\
\cellcolor[HTML]{EFEFEF}Fast Fourier -- thumb, 2 channels \cite{barron2017fast} & 2.06 & 1.39 & 0.39 & 4.80 \\
\cellcolor[HTML]{EFEFEF}Fast Fourier -- full, 4 channels \cite{barron2017fast}& 1.99 & \cellcolor[HTML]{FFFC9E}1.31 & \cellcolor[HTML]{FFFC9E}0.35 & 4.75 \\
\cellcolor[HTML]{EFEFEF}Quasi-unsupervised (tuned) \cite{bianco2019quasi} & \cellcolor[HTML]{FFFC9E}1.97 & 1.41 & - & - \\
\hline
 Avg. result for sensor-independent  & 4.26 & 3.25 & 0.99 & 9.43 \\
 Avg. result for sensor-dependent & 2.40 & 1.64 & 0.50 & 5.75\\
  \hdashline

Sensor-independent (Ours) & 2.05 & 1.50 & 0.52 & \cellcolor[HTML]{FFFC9E}4.48

\end{tabular}
}
}
\hfill
\parbox{.48\linewidth}{
\centering
\scalebox{0.578}{
\begin{tabular}{l|cccc}
\textbf{\begin{tabular}[c]{@{}l@{}}\textbf{Gehler-Shi Dataset}\\\textbf{Method} \end{tabular}} & \textbf{Mean} & \textbf{Med.} & \textbf{\begin{tabular}[c]{@{}c@{}}Best \\ 25\%\end{tabular}} & \textbf{\begin{tabular}[c]{@{}c@{}}Worst \\ 25\%\end{tabular}} \\ \hline
White-Patch \cite{maxRGB}& 7.55 & 5.68 & 1.45 & 16.12 \\
Edge-based Gamut \cite{PixelGamut}& 6.52 & 5.04 & 5.43 & 13.58 \\
Grey-world (GW) \cite{GW}& 6.36 & 6.28 & 2.33 & 10.58 \\
1st-order Gray-Edge \cite{GE}& 5.33 & 4.52 & 1.86 & 10.03 \\
2nd-order Gray-Edge \cite{GE}& 5.13 & 4.44 & 2.11 & 9.26 \\
Shades-of-Gray \cite{SoG}& 4.93 & 4.01 & 1.14 & 10.20 \\
\cellcolor[HTML]{EFEFEF}Bayesian \cite{gehler2008bayesian} & 4.82 & 3.46 & 1.26 & 10.49 \\
\cellcolor[HTML]{EFEFEF}Pixels-based Gamut \cite{PixelGamut}& 4.20 & 2.33 & 0.50 & 10.72 \\
Quasi-unsupervised \cite{bianco2019quasi} & 3.46 & 2.23 & - & - \\
PCA-based B/W Colors \cite{cheng2014illuminant}& 3.52 & 2.14 & 0.50 & 8.74 \\
\cellcolor[HTML]{EFEFEF}NetColorChecker \cite{BMVC1} & 3.10 & 2.30 & - & - \\
Grayness Index \cite{GI}& 3.07 & 1.87 & 0.43 & 7.62 \\
\cellcolor[HTML]{EFEFEF}Meta-AWB w 20 tuning images \cite{mcdonagh2018meta}& 3.00 & 2.02 & 0.58 & 7.17 \\
\cellcolor[HTML]{EFEFEF}Quasi-unsupervised \cite{bianco2019quasi} (tuned) & 2.91 &  1.98 & - & - \\
\cellcolor[HTML]{EFEFEF}Corrected-Moment \cite{MomentCorrection}& 2.86 & 2.04 & 0.70 & 6.34 \\
\cellcolor[HTML]{EFEFEF}APAP using GW \cite{Afifi19}& 2.76 & 2.02 & 0.53 & 6.21 \\
\cellcolor[HTML]{EFEFEF}Bianco et al.'s CNN \cite{bianco2015color} & 2.63 & 1.98 & 0.72 & 3.90 \\
\cellcolor[HTML]{EFEFEF}Effective Regression Tree \cite{cheng2015effective}& 2.42 & 1.65 & 0.38 & 5.87 \\
\cellcolor[HTML]{EFEFEF}WB-sRGB (modified for raw-RGB) \cite{afifi2019color}& 2.07 & 1.38 &  \cellcolor[HTML]{FFFC9E}0.29 &
5.11 \\
\cellcolor[HTML]{EFEFEF}Fast Fourier - thumb, 2 channels \cite{barron2017fast}& 2.01 & 1.13 & 0.30 & 5.14 \\
\cellcolor[HTML]{EFEFEF}Conv Color Constancy \cite{barron2015convolutional}& 1.95 & 1.22 & 0.35 & 4.76 \\
\cellcolor[HTML]{EFEFEF}Deep Specialized Net \cite{shi2016deep}& 1.90 & 1.12 & 0.31 & 4.84 \\
\cellcolor[HTML]{EFEFEF}Fast Fourier - full, 4 channels \cite{barron2017fast}& 1.78 & \cellcolor[HTML]{FFFC9E}0.96 & \cellcolor[HTML]{FFFC9E}0.29 & 4.62 \\
\cellcolor[HTML]{EFEFEF}AlexNet-FC4 \cite{hu2017fc}& 1.77 & 1.11 & 0.34 & 4.29 \\
\cellcolor[HTML]{EFEFEF}SqueezeNet-FC4 \cite{hu2017fc}& \cellcolor[HTML]{FFFC9E}1.65 & 1.18 & 0.38 & \cellcolor[HTML]{FFFC9E}3.78 \\
\hline
 Avg. result for sensor-independent  & 5.10 & 4.03 & 1.91 & 10.77\\
 Avg. result for sensor-dependent & 2.62 & 1.75  & 0.50 & 5.95\\
 \hdashline
Sensor-independent (Ours) & 2.77 & 1.93 & 0.55 & 6.53
\end{tabular}}
}\vspace{-3mm}
\end{table}

In our experiments, we used all cameras of three different datasets, which are: (i) NUS 8-Camera \cite{cheng2014illuminant}, (ii) Gehler-Shi \cite{gehler2008bayesian}, and (iii) Cube+ \cite{banic2017unsupervised} datasets. In total, we have 4,014 raw-RGB images captured by 11 different camera sensors.

We followed the leave-one-out cross-validation scheme to evaluate our method. Specifically, we excluded all images captured by one camera for testing and trained a model with the remaining images. This process was repeated for all cameras. We also tested our method on the Cube dataset. In this experiment, we used a trained model on images from the NUS and Gehler-Shi datasets, and excluded all images from the Cube+ dataset. The calibration objects (i.e., X-Rite color chart or SpyderCUBE) were masked out in both training and testing processes.

\begin{figure}
\begin{center}
\includegraphics[width=0.97\linewidth]{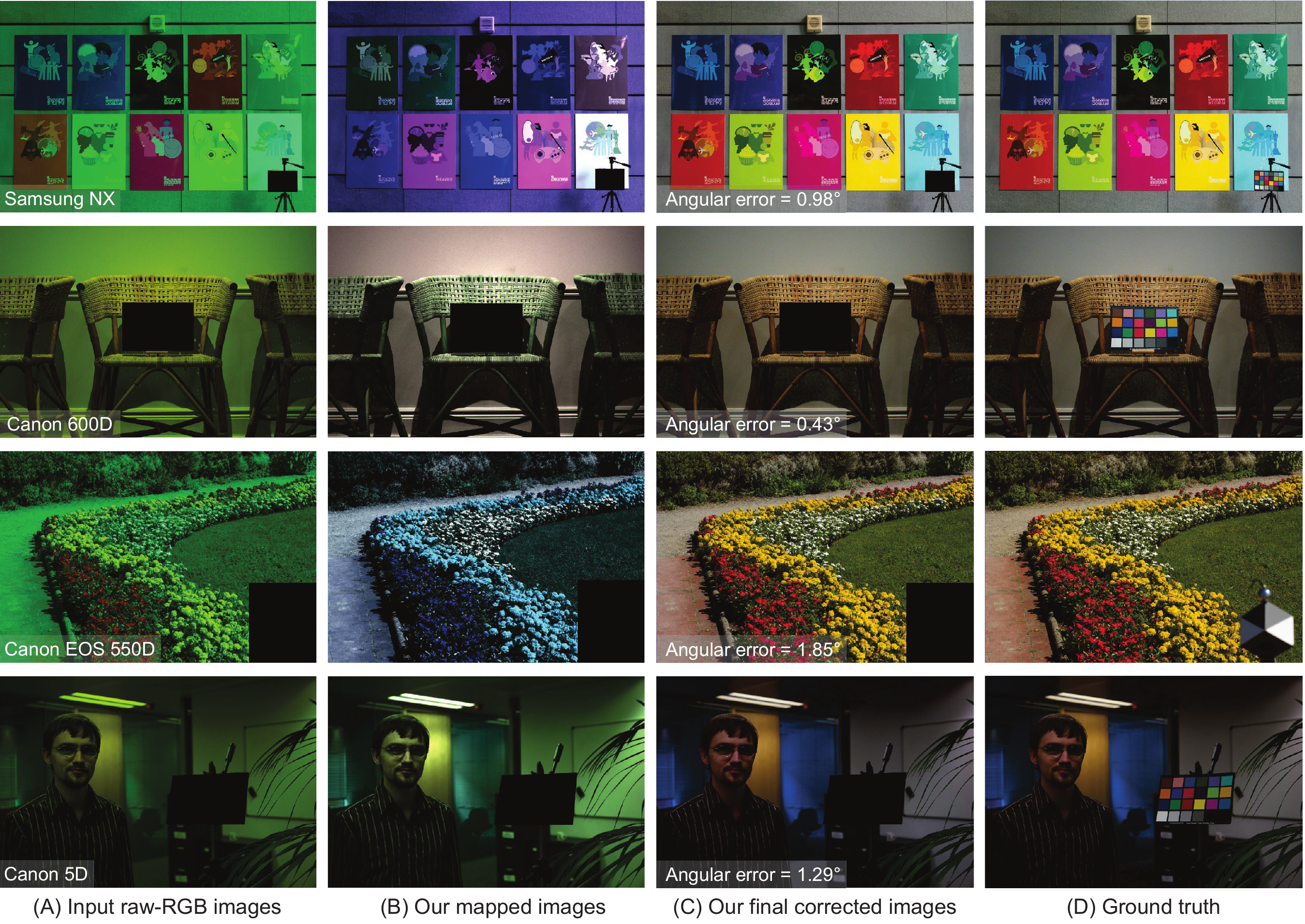}
\end{center}
\vspace{-5mm}
\caption{Qualitative results of our method. (A) Input raw-RGB images. (B) After mapping images in (A) to the learned space. (C) After correcting images in (A) based on our estimated illuminants. (D) Corrected by ground truth illuminants. Shown images are rendered in the sRGB color space by the camera imaging pipeline in \cite{karaimer2016software} to aid visualization.}\vspace{-5mm}
\label{fig:results}
\end{figure}

Unlike results reported by existing learning methods which use three-fold cross-validation for evaluation, our reported results were obtained by models that were {\it not} trained on any example of the testing camera sensor.

In Tables \ref{Table1}--\ref{Table2}, the mean, median, best 25\%, and the worst 25\% of the angular error between our estimated illuminants and ground truth are reported. The best 25\% and worst 25\% are the mean of the smallest 25\% angular error values and the mean of the highest 25\% angular error
values, respectively. We highlight learning methods (i.e., models trained/tuned for the testing sensor) with gray in the shown tables. The reported results are taken from previous papers, except for the recent work in \cite{afifi2019color}, which was proposed for white balancing images saved in the sRGB color space. We modified \cite{afifi2019color} to work in the raw-RGB space by replacing the training polynomial matrices with the ground truth illuminant vectors. The shown results of \cite{afifi2019color} were obtained by using training data from a single camera sensor (i.e., sensor-specific) with the following settings: $k=15$, $\sigma=0.45$, $m=91$, and $c=191$. For the recent work in \cite{bianco2019quasi}, we include results of the unsupervised and tuned models. Our method performs better than all statistical-based methods and outperforms some sensor-specific learning methods. We obtain results on par with the \textit{sensor-specific} state-of-the-art results in the NUS 8-Camera dataset (Table \ref{Table1}).  

\begin{table}
\caption{Angular errors on the Cube and Cube+ datasets \cite{banic2017unsupervised}. Methods highlighted in gray are trained/tuned for each camera sensor (i.e., sensor-specific models). The lowest errors are highlighted in yellow. \label{Table2}}\vspace{2mm}
\parbox{.48\linewidth}{
\centering
\scalebox{0.59}
{
\begin{tabular}{l|cccc}
\textbf{\begin{tabular}[c]{@{}l@{}}\textbf{Cube Dataset}\\\textbf{Method} \end{tabular}} & \textbf{Mean} & \textbf{Med.} & \textbf{\begin{tabular}[c]{@{}c@{}}Best \\ 25\%\end{tabular}} & \textbf{\begin{tabular}[c]{@{}c@{}}Worst \\ 25\%\end{tabular}} \\ \hline
White-Patch \cite{maxRGB} & 6.58 & 4.48 & 1.18 & 15.23 \\
Grey-world (GW) \cite{GW} & 3.75 & 2.91 & 0.69 & 8.18 \\
Shades-of-Gray \cite{SoG} & 2.58 & 1.79 & 0.38 & 6.19 \\
2nd-order Gray-Edge \cite{GE}& 2.49 & 1.60 & 0.49 & 6.00 \\
1st-order Gray-Edge \cite{GE}& 2.45 & 1.58 & 0.48 & 5.89 \\

\cellcolor[HTML]{EFEFEF}APAP using GW \cite{Afifi19}& 1.55  & 1.02 & 0.28 &  3.74 \\

\cellcolor[HTML]{EFEFEF}Color Dog \cite{banic2015color}&  1.50 & 0.81 &  0.27 & 3.86 \\
\cellcolor[HTML]{EFEFEF}Meta-AWB (20) \cite{mcdonagh2018meta}& 1.74 & 1.08 & 0.29 & 4.28 \\
\cellcolor[HTML]{EFEFEF}WB-sRGB (modified for raw-RGB) \cite{afifi2019color}& \cellcolor[HTML]{FFFC9E} 1.37 & \cellcolor[HTML]{FFFC9E}0.78 & \cellcolor[HTML]{FFFC9E}0.19 &
\cellcolor[HTML]{FFFC9E}3.51 \\
\hline
 Avg. result for sensor-independent  &  3.57 & 2.47 & 0.64 & 8.30\\
 Avg. result for sensor-dependent & 1.54 & 0.92 & 0.26 & 3.85\\
  \hdashline
Sensor-independent (Ours) & 1.98 & 1.36 & 0.40 & 4.64

\end{tabular}
}
}
\hfill
\parbox{.48\linewidth}{
\centering
\scalebox{0.575}
{
\begin{tabular}{l|cccc}
\textbf{\begin{tabular}[c]{@{}l@{}}\textbf{Cube+ Dataset}\\\textbf{Method} \end{tabular}} & \textbf{Mean} & \textbf{Med.} & \textbf{\begin{tabular}[c]{@{}c@{}}Best \\ 25\%\end{tabular}} & \textbf{\begin{tabular}[c]{@{}c@{}}Worst \\ 25\%\end{tabular}} \\ \hline
White-Patch \cite{maxRGB}& 9.69 & 7.48 & 1.72 & 20.49 \\
Grey-world (GW) \cite{GW}& 7.71 & 4.29 & 1.01 & 20.19 \\
\cellcolor[HTML]{EFEFEF}Color Dog \cite{banic2015color}& 3.32 & 1.19 & 0.22 & 10.22 \\
Shades-of-Gray \cite{SoG}& 2.59 & 1.73 & 0.46 & 6.19 \\
2nd-order Gray-Edge \cite{GE}& 2.50 & 1.59 & 0.48 & 6.08 \\
1st-order Gray-Edge \cite{GE}& 2.41 & 1.52 & 0.45 & 5.89 \\
\cellcolor[HTML]{EFEFEF}APAP using GW \cite{Afifi19}& 2.01  &  1.36 & 0.38 & 4.71 \\
\cellcolor[HTML]{EFEFEF}Color Beaver \cite{kovsvcevic2019color}& 1.49 & 0.77 & 0.21 & 3.94 \\

\cellcolor[HTML]{EFEFEF}WB-sRGB (modified for raw-RGB) \cite{afifi2019color}& \cellcolor[HTML]{FFFC9E} 1.32 & \cellcolor[HTML]{FFFC9E}0.74 & \cellcolor[HTML]{FFFC9E}0.18 &
\cellcolor[HTML]{FFFC9E}3.43 \\

\hline
 Avg. result for sensor-independent  & 4.98 & 3.32 & 0.82 & 11.77\\

 Avg. result for sensor-dependent & 2.04 & 1.02 & 0.25 & 5.58\\
  \hdashline
Sensor-independent (Ours) & 2.14 & 1.44 & 0.44 &  5.06
\end{tabular}

}
}\vspace{-3mm}
\end{table}

\begin{table}
\centering
\caption{Angular and reproduction angular errors \cite{finlayson2016reproduction} on the Cube+ challenge \cite{challenge}. The methods are sorted by the median of the errors (shown in bold), as ranked in the challenge \cite{challenge}. Methods highlighted in gray are sensor-specific models. We show our results w/wo training on Cube+ dataset. The lowest errors over all methods are highlighted in yellow. \label{Table3}}\vspace{2mm}
\parbox{.48\linewidth}{
\scalebox{0.578}
{
\begin{tabular}{l|cccc}
\textbf{\begin{tabular}[c]{@{}l@{}}\textbf{Cube+ challenge (angular error)}\\\textbf{Method} \end{tabular}} & \textbf{Mean} & \textbf{Med.} & \textbf{\begin{tabular}[c]{@{}c@{}}Best \\ 25\%\end{tabular}} & \textbf{\begin{tabular}[c]{@{}c@{}}Worst \\ 25\%\end{tabular}} \\ \hline

Grey-world (GW) \cite{GW} & 4.44 & \textbf{3.50} & 0.77 & 9.64 \\
1st-order Gray-Edge \cite{GE} & 3.51 & \textbf{2.3} & 0.56 & 8.53 \\
 V Vuk et al., \cite{challenge}& 6 & \textbf{1.96} & 0.99 & 18.81 \\
\cellcolor[HTML]{EFEFEF} Y Qian et al., (1) \cite{challenge}& 2.48 & \textbf{1.56} & 0.44 & 6.11 \\
\cellcolor[HTML]{EFEFEF} Y Qian et al., (2) \cite{challenge}& 1.84 & \textbf{1.27} & 0.39 & 4.41 \\
\cellcolor[HTML]{EFEFEF} K Chen et al., \cite{challenge}& 1.84 & \textbf{1.27} & 0.39 & 4.41 \\
\cellcolor[HTML]{EFEFEF} Y Qian et al., (3) \cite{challenge}& 2.27 & \textbf{1.26} & 0.39 & 6.02 \\
\cellcolor[HTML]{EFEFEF} Fast Fourier \cite{barron2017fast}& 2.1 & \textbf{1.23} & 0.47 & 5.38\\
\cellcolor[HTML]{EFEFEF} A Savchik et al., \cite{challenge}& 2.05 & \textbf{1.2} & 0.41 & 5.24 \\
\cellcolor[HTML]{EFEFEF} WB-sRGB (modified for raw-RGB) \cite{afifi2019color} & \cellcolor[HTML]{FFFC9E} 1.83 & \cellcolor[HTML]{FFFC9E} \textbf{1.15} & \cellcolor[HTML]{FFFC9E} 0.35 & \cellcolor[HTML]{FFFC9E} 4.6\\
\hdashline
Ours trained wo/ Cube+ & 2.89 & \textbf{1.72} & 0.71 & 7.06 \\
\cellcolor[HTML]{EFEFEF} Ours trained w/ Cube+ & 2.1 & \textbf{1.23} & 0.47 & 5.38
\end{tabular}
}
}
\hfill
\parbox{.48\linewidth}{
\scalebox{0.578}
{
\begin{tabular}{l|cccc}
\textbf{\begin{tabular}[c]{@{}l@{}}\textbf{Cube+ challenge (reproduction error)}\\\textbf{Method} \end{tabular}} & \textbf{Mean} & \textbf{Med.} & \textbf{\begin{tabular}[c]{@{}c@{}}Best \\ 25\%\end{tabular}} & \textbf{\begin{tabular}[c]{@{}c@{}}Worst \\ 25\%\end{tabular}} \\ \hline

Grey-world (GW) \cite{GW} & 5.74 & \textbf{4.60} & 1.12 & 12.21 \\
1st-order Gray-Edge \cite{GE} & 4.57 & \textbf{3.22} & 0.84 & 10.75 \\
V Vuk et al., \cite{challenge}& 6.87 & \textbf{2.1} & 1.06 & 21.82 \\
\cellcolor[HTML]{EFEFEF} Y Qian et al., (1) \cite{challenge}& 6.87 & \textbf{2.09} & 0.61 & 8.18 \\
\cellcolor[HTML]{EFEFEF} Y Qian et al., (2) \cite{challenge}& 2.49 & \textbf{1.71} & 0.52 & 6 \\
\cellcolor[HTML]{EFEFEF} K Chen et al., \cite{challenge}& 2.49 & \textbf{1.69} & 0.52 & 6 \\
\cellcolor[HTML]{EFEFEF} Y Qian et al., (3) \cite{challenge}& 2.93 & \textbf{1.64} & 0.5 & 7.78 \\
\cellcolor[HTML]{EFEFEF} Fast Fourier \cite{barron2017fast}& 2.48 & \textbf{1.59} & 0.58 & 7.27 \\
\cellcolor[HTML]{EFEFEF} A Savchik et al., \cite{challenge}& 2.65 & \textbf{1.51} & 0.5 & 6.85 \\
\cellcolor[HTML]{EFEFEF} WB-sRGB (modified for raw-RGB) \cite{afifi2019color} & \cellcolor[HTML]{FFFC9E} 2.36 & \cellcolor[HTML]{FFFC9E} \textbf{1.47} & \cellcolor[HTML]{FFFC9E} 0.45 & \cellcolor[HTML]{FFFC9E} 5.94\\

\hdashline
Ours trained wo/ Cube+ & 3.97 & \textbf{2.31} & 0.86 & 10.07\\
\cellcolor[HTML]{EFEFEF} Ours trained w/ Cube+ & 2.8 & \textbf{1.54} & 0.58 &  7.27
\end{tabular}

}
}
\end{table}

We also examined our trained models on the Cube+ challenge \cite{challenge}. This challenge introduced a new testing set of 363 raw-RGB images captured by a Canon EOS 550 D -- the same camera model used in the original Cube+ dataset \cite{banic2017unsupervised}. In our results, we did not include any image from the testing set in the training/validation processes. Instead, we used the same models trained for the evaluation on the other datasets (Tables \ref{Table1}--\ref{Table2}). Table \ref{Table3} shows the angular error and reproduction angular errors \cite{finlayson2016reproduction} obtained by our models and the top-ranked methods that participated in the challenge. Additionally, we show results obtained by other methods \cite{GW, SoG, afifi2019color}. For \cite{afifi2019color}, we show the results of the ensemble model (i.e., averaged estimated illuminant vectors from the three-fold trained models on the Cube+ dataset). We report results of two trained models using our method. The first one was trained without examples from Cube+ camera sensor (i.e., trained on all camera models in NUS and Gehler-Shi datasets). The second model was originally trained to evaluate our method on one camera of the NUS 8-Cameras dataset (i.e., trained on seven out of the eight camera models in NUS 8-Cameras dataset, the Cube+ camera model, and the Gehler-Shi camera models). The latter model is provided to demonstrate the ability of our method to use different camera models beside the target camera model during the training phase. More results of the Cube+ challenge are provided in the supplemental materials.

We further tested our trained models on the INTEL-TUT dataset \cite{aytekin2018data}, which includes DSLR and mobile phone cameras that are not included in the NUS 8-Camera, Gehler-Shi, and Cube+ datasets. Table \ref{Table4} shows the obtained results by the proposed method trained on DSLR cameras from the NUS 8-Camera, Gehler-Shi, and Cube+ datasets.

Finally, we show qualitative examples in Fig. \ref{fig:results}. For each example, we show the mapped image $\mathbf{I}_m$ in our learned intermediate space. In the shown figure, we rendered the images in the sRGB color space by the camera imaging pipeline in \cite{karaimer2016software} to aid visualization.

\begin{table}[]
\centering
\caption{Angular errors on the INTEL-TUT dataset \cite{aytekin2018data}. Methods highlighted in gray are trained/tuned for each camera sensor (i.e., sensor-specific models). The lowest errors are highlighted in yellow. \label{Table4}}\vspace{2mm}
\scalebox{0.48}
{
\begin{tabular}{c|ccccccc:cc}
\textbf{\begin{tabular}[c]{@{}l@{}}\textbf{INTEL-TUT Dataset} \end{tabular}} & Gray-World \cite{GW}& \begin{tabular}[c]{@{}c@{}}Shades-\\of-Gray \cite{SoG}\end{tabular}& \begin{tabular}[c]{@{}c@{}}2nd-order\\  Gray-Edge \cite{GE}\end{tabular} & \begin{tabular}[c]{@{}c@{}}PCA-based \\ B/W Colors \cite{cheng2014illuminant}\end{tabular} & \begin{tabular}[c]{@{}c@{}}1st-order \\ Gray-Edge \cite{GE}\end{tabular} & \cellcolor[HTML]{EFEFEF}\begin{tabular}[c]{@{}c@{}}
APAP\\ using\\ GW \cite{Afifi19}\end{tabular}& \cellcolor[HTML]{EFEFEF}\begin{tabular}[c]{@{}c@{}}
WB-sRGB \cite{afifi2019color}\\ (modified for raw-RGB) \end{tabular} &  \begin{tabular}[c]{@{}c@{}}Ours \\trained on NUS \\ and Cube+\end{tabular} & \begin{tabular}[c]{@{}c@{}}Ours\\ trained on NUS\\  and Gehler-Shi\end{tabular} \\ \hline
\textbf{Mean} & 4.77 & 4.99 & 4.82 & 4.65 & 4.62 & 4.30 & \cellcolor[HTML]{FFFC9E} 1.79 & 3.76 & 3.82 \\
\textbf{Median} & 3.75 & 3.63 & 2.97 & 3.39 & 2.84 & 2.44 & \cellcolor[HTML]{FFFC9E} 0.87 & 2.75 &  2.81 \\
\textbf{Best 25\%} & 0.99 & 1.08 & 1.03 & 0.87 & 0.94 & 0.69 & \cellcolor[HTML]{FFFC9E} 0.14 & 0.81 & 0.87 \\
\textbf{Worst 25\% }& 10.29 & 11.20 & 11.96 & 10.75 & 11.46 & 11.30 & \cellcolor[HTML]{FFFC9E} 5.08 & 8.40 & 8.65
\end{tabular}
}\vspace{-3mm}
\end{table}

\section{Conclusion}
\label{sec:conclusion}
We have proposed a deep learning method for illuminant estimation. Unlike other learning-based methods, our method is a sensor-independent and can be trained on images captured by different camera sensors. To that end, we have introduced an image-specific learnable mapping matrix that maps an input image to a new sensor-independent space. Our method relies only on color distributions of images to estimate scene illuminants. We adopted a compact color histogram that is dynamically generated by our new RGB-$uv$ histogram block. Our method achieves good results on images captured by new camera sensors that have not been used in the training process.

\paragraph{Acknowledgment}  This study was funded in part by the Canada First Research Excellence Fund for the Vision: Science to Applications (VISTA) programme and an NSERC Discovery Grant.  Dr. Brown contributed to this article in his personal capacity as a professor at York University.  The views expressed (or the conclusions reached) are his own and do not necessarily represent the views of Samsung Research.

\end{document}